\documentclass[runningheads]{llncs}
\usepackage[T1]{fontenc}

\usepackage{cite}
\usepackage{amsmath,amssymb,amsfonts}
\usepackage{algorithmic}
\usepackage{graphicx}
\usepackage{textcomp}
\usepackage{xcolor}
\usepackage{booktabs}
\usepackage{array}
\usepackage{multirow}
\usepackage{tikz}
\usetikzlibrary{arrows.meta, shapes.geometric, positioning, fit, backgrounds, decorations.pathreplacing}
\usepackage{pgfplots}
\pgfplotsset{compat=1.18}
\usepackage{hyperref}
\usepackage{url}
\usepackage{float}
\usepackage{comment}

\def\BibTeX{{\rm B\kern-.05em{\sc i\kern-.025em b}\kern-.08em
    T\kern-.1667em\lower.7ex\hbox{E}\kern-.125emX}}

\begin{document}

\titlerunning{Quantum Kernel Based Classification for Quishing Detection}

\title{Quantum Kernel Based Classification for Quishing Detection:
A Comparative Study of QKNN and QSVM}

\author{Md Farhan Shahriyar\inst{1}$^{*}$}

\authorrunning{Md Farhan Shahriyar}

\institute{
  Department of Computer Science and Engineering,\\
  Hajee Mohammad Danesh Science and Technology University,\\
  Dinajpur , Bangladesh\\
  \email{farhanshahriyar.@gmail.com}\\[4pt]
  $^{*}$Corresponding author
}

\maketitle

\begin{abstract}
Quishing , QR code based phishing poses a rapidly growing threat in modern cybersecurity yet its automated detection via machine learning remains underexplored. This paper presents a comprehensive empirical evaluation of Quantum K-Nearest Neighbors (QKNN) and Quantum Support Vector Machines (QSVM/QKSVM) for binary classification of benign versus malicious QR code images. Implemented using PennyLane on a balanced dataset of  QR code images, we systematically compare four quantum kernel variants Angle Embedding, ZZ Feature Map, IQP Embedding, and Amplitude-Entangling against classical SVM baselines with linear, RBF and polynomial kernels. Additionally, we evaluate Hybrid Quantum-Classical Neural Networks (HQCNN) in which quantum layers replace classical processing blocks and further examine Quantum-Classical skip-connection architectures. Experiments employ three fold and five fold stratified cross validation protocols. Results demonstrate that the trainable quantum fidelity kernel employed in QKNN achieves a superior AUC of 0.7601 versus the classical KNN AUC of 0.7449 at K=15 consistently outperforming classical counterparts in ranking quality. A detailed methodology diagram and ablation studies on quantum circuit depth and PCA dimensionality are also provided contributing a reproducible framework for quantum machine learning in cybersecurity applications.

\keywords{Quantum Machine Learning\and Quantum Kernel \and QKNN\and QSVM\and Quishing Detection\and QR Code Security\and Hybrid Quantum-Classical \and PennyLane\and ZZFeatureMap\and Variational Quantum Circuit}

\end{abstract}
\section{Introduction}

QR codes have become ubiquitous in digital communication, enabling seamless redirection to websites, payment portals, and authentication services. Their widespread adoption has simultaneously created an attack surface for cybercriminals: the called quishing attacks embed malicious URLs within QR codes bypassing traditional URL-based phishing filters that operate on plaintext hyperlinks \cite{quishing_ml,shahriyar2025phishvqc}. Unlike conventional phishing, quishing exploits the opacity of encoded payloads visual scanners are required to decode the embedded content, making automated detection nontrivial.

Classical machine learning approaches have been applied to QR code classification with moderate success; however, high-dimensional image feature spaces and the subtle visual similarity between benign and malicious QR codes continue to challenge standard linear and kernel methods. Quantum computing offers a compelling alternative paradigm: quantum kernel functions can map classical data into exponentially large Hilbert spaces, potentially revealing separating hyperplanes inaccessible to classical kernels \cite{schuld_kernel}. Similarly, variational quantum circuits can serve as expressive feature extractors within hybrid deep learning pipelines.

Despite growing interest in Quantum Machine Learning (QML), direct comparative studies of QKNN and QSVM on cybersecurity image datasets remain scarce\cite{shahriyar2025toward}. Existing work has demonstrated quantum kernel advantages in controlled benchmarks \cite{havlicek_qsvm}, but end-to-end evaluations on realistic threat-detection tasks, with rigorous cross-validation and ablation studies, are lacking. This paper addresses this gap with the following contributions:

\begin{enumerate}
    \item A systematic comparison of four quantum kernel variants (Angle, ZZ, IQP, Amplitude-Entangling) against three classical SVM kernels on a QR code quishing dataset
    \item A trainable quantum fidelity kernel for QKNN optimized via Kernel-Target Alignment (KTA) with comprehensive K-value sweep, circuit-depth ablation and PCA-dimension ablation studies.

\end{enumerate}

The remainder of this paper is organized as follows. Section~\ref{sec:related} reviews related work. Section~\ref{sec:dataset} describes the dataset. Section~\ref{sec:method} presents the methodology in detail. Section~\ref{sec:results} presents experimental results. Section~\ref{sec:discussion} discusses key findings, and Section~\ref{sec:conclusion} concludes.

\section{Related Work}
\label{sec:related}

\subsection{Quantum Kernel Methods and QSVM}

Support Vector Machines occupy a central role in classical machine learning by implicitly operating in high-dimensional feature spaces via the kernel trick \cite{cortes_svm}. The quantum analogue replaces classical kernels with fidelity-based quantum kernels: given data encoding circuits \(\mathcal{U}(\mathbf{x})\), the quantum kernel evaluates \( k(\mathbf{x}_i, \mathbf{x}_j) = |\langle 0 | \mathcal{U}^\dagger(\mathbf{x}_j) \mathcal{U}(\mathbf{x}_i) | 0 \rangle|^2 \). The study  \cite{havlicek_qsvm} demonstrated that ZZ Feature Map-based quantum kernels can, in principle, achieve classification advantages over classical methods for datasets with specific quantum structure. Subsequent work explored trainable quantum kernels optimized by Kernel-Target Alignment \cite{kta_training}, IQP embedding circuits \cite{iqp_hardness}, and amplitude-entangling strategies. For image classification tasks, hybrid pipelines using ResNet feature extractors followed by QSVM have demonstrated competitive performance on disease detection benchmarks \cite{shahriyar2025quantum}, suggesting that the quantum kernel layer benefits from receiving compressed, semantically rich feature representations rather than raw pixel inputs.

Classical KNN classifies query points by plurality vote among the \(k\) nearest training samples, with distance measured in Euclidean or other metric spaces. Quantum KNN (QKNN) replaces the distance computation with quantum state overlap \cite{qknn_swap, qknn_mahalanobis}, leveraging the SWAP test or fidelity measurement to quantify similarity in Hilbert space. This approach can in theory evaluate distances between exponentially high-dimensional quantum state representations efficiently. Recent implementations of QKNN have reported accuracy improvements in printer forensics \cite{qknn_printer} and benchmark classification tasks \cite{qknn_noise_resilient}. A quantum circuit-based KNN supporting simultaneous \(K\) value selection was formalized in \cite{qknn_ieee_circuit}. Notably, the noise-resilient QKNN of \cite{qknn_noise_resilient} achieved up to 100\% accuracy on controlled datasets, suggesting strong potential for practical deployment as quantum hardware matures.

Hybrid Quantum-Classical Convolutional Neural Networks (HQCNN) embed variational quantum circuits as drop-in replacements for classical neural network layers \cite{hqcnn_layerwise, schuld_vqc,tanbhir2025quanvolutional}. The quantum sublayer receives encoded classical activations, applies parameterized rotation and entangling gates, and returns expectation values as the next layer's input. Systematic studies of quantum layer placement early, middle, late, or at the classifier head reveal that mid-level quantum substitution often balances expressivity and trainability \cite{hqcnn_placement}, echoing findings in classical architecture search.
Quishing detection has primarily relied on URL analysis and content-based filtering \cite{quishing_ml}. Image-level ML approaches treating QR codes as visual patterns are nascent, with binary classification (benign/malicious) being a natural formulation given the two-class nature of most threat models. To the authors' knowledge, this is the first study applying both QKNN and QSVM with multiple quantum kernels to QR code quishing classification.

\section{Methodology}
\label{sec:method}

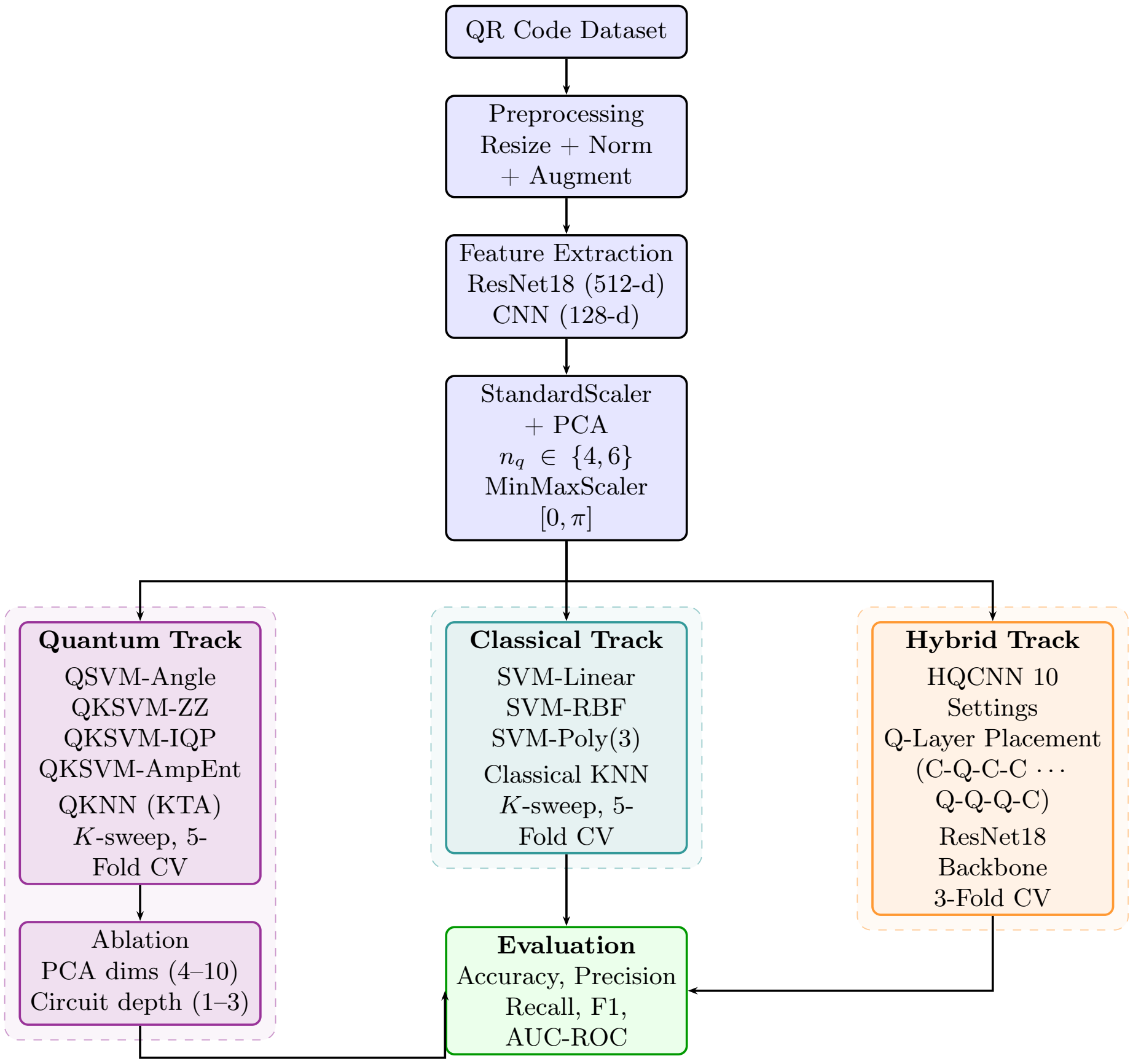
\begin{figure}[t]
\centering
\resizebox{\columnwidth}{!}{%
\begin{tikzpicture}[
  node distance=0.45cm,
  every node/.style={font=\small},
  box/.style={
    rectangle, rounded corners=3pt, draw=black, thick,
    fill=blue!10, text width=2.8cm,
    minimum height=0.65cm, align=center},
  qbox/.style={
    rectangle, rounded corners=3pt, draw=violet!80, thick,
    fill=violet!12, text width=2.8cm,
    minimum height=0.70cm, align=center},
  cbox/.style={
    rectangle, rounded corners=3pt, draw=teal!80, thick,
    fill=teal!10, text width=2.8cm,
    minimum height=0.70cm, align=center},
  hbox/.style={
    rectangle, rounded corners=3pt, draw=orange!80, thick,
    fill=orange!10, text width=2.8cm,
    minimum height=0.70cm, align=center},
  rbox/.style={
    rectangle, rounded corners=3pt, draw=green!60!black, thick,
    fill=green!8, text width=2.8cm,
    minimum height=0.65cm, align=center},
  arr/.style={-{Stealth[length=3.5pt]}, thick},
]

\node[box] (data)   {QR Code Dataset};
\node[box, below=of data]  (prep) {Preprocessing\\Resize + Norm + Augment};
\node[box, below=of prep]  (feat) {Feature Extraction\\ResNet18 (512-d)\\CNN (128-d)};
\node[box, below=of feat]  (pca)  {StandardScaler + PCA\\$n_q \in \{4,6\}$\\MinMaxScaler $[0,\pi]$};

\draw[arr] (data)--(prep);
\draw[arr] (prep)--(feat);
\draw[arr] (feat)--(pca);

\node[qbox, below left=1.0cm and 2.3cm of pca] (qt)
    {\textbf{Quantum Track}\\[2pt]
     QSVM-Angle\\QKSVM-ZZ\\QKSVM-IQP\\QKSVM-AmpEnt\\[2pt]
     QKNN (KTA)\\$K$-sweep, 5-Fold CV};

\node[cbox, below=1.0cm of pca] (ct)
    {\textbf{Classical Track}\\[2pt]
     SVM-Linear\\SVM-RBF\\SVM-Poly(3)\\[2pt]
     Classical KNN\\$K$-sweep, 5-Fold CV};

\node[hbox, below right=1.0cm and 2.3cm of pca] (ht)
    {\textbf{Hybrid Track}\\[2pt]
     HQCNN 10 Settings\\Q-Layer Placement\\(C-Q-C-C $\cdots$ Q-Q-Q-C)\\[2pt]
     ResNet18 Backbone\\3-Fold CV};

\draw[arr] (pca.south) -- ++(0,-0.5) -| (qt.north);
\draw[arr] (pca.south) --              (ct.north);
\draw[arr] (pca.south) -- ++(0,-0.5) -| (ht.north);

\node[qbox, below=0.45cm of qt] (abl)
    {Ablation\\PCA dims (4--10)\\Circuit depth (1--3)};

\draw[arr] (qt)--(abl);

\node[rbox, below=0.9cm of ct] (eval)
    {\textbf{Evaluation}\\Accuracy, Precision\\Recall, F1, AUC-ROC};

\draw[arr] (ct.south)  -- (eval.north);
\draw[arr] (ht.south)  |- (eval.east);

\draw[arr] (abl.south)
    -- ++(0,-0.4)
    -| (eval.west);

\begin{scope}[on background layer]
  \node[draw=violet!40, dashed, rounded corners=5pt,
        fill=violet!4, inner sep=5pt, fit=(qt)(abl)] {};
  \node[draw=teal!40,   dashed, rounded corners=5pt,
        fill=teal!4,   inner sep=5pt, fit=(ct)]      {};
  \node[draw=orange!40, dashed, rounded corners=5pt,
        fill=orange!4, inner sep=5pt, fit=(ht)]      {};
\end{scope}

\end{tikzpicture}
}%
\caption{Methodology of the proposed quantum-classical framework for QR code quishing detection.}

\label{fig:methodology}
\end{figure}

Fig.~\ref{fig:methodology} illustrates the complete experimental pipeline. The pipeline branches into three main tracks: (1) QSVM/classical SVM evaluation, (2) QKNN evaluation, and (3) HQCNN evaluation.

\subsection{Dataset Description}
\label{sec:dataset}

The dataset consists of  QR code images benign and  malicious constituting a perfectly balanced binary classification benchmark for quishing detection\cite{galadima2025qr}. For the QSVM experiments, images were resized to 64\(\times\)64 pixels and converted to grayscale. For the HQCNN experiments, images were resized to 224\(\times\)224 pixels (RGB) following ResNet18 preprocessing standards. Data augmentation during training included random horizontal flip, affine transformations (\(\pm 5°\) rotation, \(\pm 5\%\) translation), and color jitter (brightness and contrast \(\pm 0.2\)). Normalization followed ImageNet statistics (\(\mu = [0.485, 0.456, 0.406]\), \(\sigma = [0.229, 0.224, 0.225]\)). For quantum model inputs, features were standardized via \texttt{StandardScaler}, reduced by PCA to \(n_{\mathrm{qubits}}\) components (4 or 6), and then scaled to \([0, \pi]\) via \texttt{MinMaxScaler} to enable angle-based quantum encoding. The PCA variance retained varied from 49.5\% (4 components) to 68.7\% (10 components) on the flattened pixel features.

\subsection{Quantum Kernel Definitions}

All quantum circuits were implemented with PennyLane 0.44.1 using the \texttt{default.qubit} simulator. Four quantum kernel functions were evaluated:

Angle Embedding Kernel (QSVM-Angle): The simplest kernel, employing \texttt{qml.AngleEmbedding} with RY rotations. For data points \(\mathbf{x}_1, \mathbf{x}_2 \in \mathbb{R}^{n_q}\), the kernel is:
\[
k_{\text{angle}}(\mathbf{x}_1, \mathbf{x}_2) = |\langle 0^n | U_{\text{RY}}^\dagger(\mathbf{x}_2)\, U_{\text{RY}}(\mathbf{x}_1) | 0^n \rangle|^2
\]

ZZ Feature Map Kernel (QKSVM-ZZ): Inspired by \cite{havlicek_qsvm}, this kernel applies Hadamard gates, single-qubit RZ rotations, and nearest-neighbor CNOT entanglement with cross-feature ZZ interaction terms \((x_i \cdot x_{i+1})\). The ZZ kernel is more expressive than angle embedding due to entanglement-induced correlations.

IQP Embedding Kernel (QKSVM-IQP): Employs \texttt{qml.IQPEmbedding} with 2 repetitions, instantiating an Instantaneous Quantum Polynomial circuit structure \cite{iqp_hardness} resistant to classical simulation.

Amplitude-Entangling Kernel (QKSVM-AmpEnt): The most expressive kernel, combining \texttt{AngleEmbedding} in both RY and RZ bases interleaved with CNOT entangling layers, constructing a deeply entangled feature map via adjoint unitary inversion.

All kernels compute the probability of measuring the all-zero bitstring, yielding a scalar fidelity \(k \in [0, 1]\). Full kernel matrices \(\mathbf{K}_{ij} = k(\mathbf{x}_i, \mathbf{x}_j)\) were precomputed.

\subsection{Trainable Quantum Kernel for QKNN}

For QKNN, a trainable ZZ-style fidelity kernel was employed. The kernel circuit consists of \(n_{\text{layers}}\) repetitions of: Hadamard layer, single-qubit RZ+RY encoding, and nearest-neighbor ZZ entangling gates. Kernel parameters \(\boldsymbol{\theta}\) were optimized by maximizing the Kernel-Target Alignment (KTA) \cite{kta_training}:
\[
\mathrm{KTA}(\mathbf{K}, \mathbf{y}) = \frac{\langle \mathbf{K}, \mathbf{y}\mathbf{y}^\top \rangle_F}{\sqrt{\langle \mathbf{K}, \mathbf{K} \rangle_F \cdot \langle \mathbf{y}\mathbf{y}^\top, \mathbf{y}\mathbf{y}^\top \rangle_F}}
\]
where labels \(\mathbf{y} \in \{-1, +1\}^N\). KTA optimization was performed via gradient descent (Adam optimizer) over a subsample of training data.

\subsection{QKNN Classification Procedure}

Given trained kernel parameters, the quantum distance matrix was derived as:
\[
D_{ij} = \sqrt{2 - 2\,k_\theta(\mathbf{x}_i, \mathbf{x}_j)}
\]
This ensures \(D_{ij} = 0\) when quantum states are identical and \(D_{ij} = \sqrt{2}\) when orthogonal. Scikit-learn's \texttt{KNeighborsClassifier} with \texttt{metric='precomputed'} and \texttt{weights='distance'} was applied to the distance matrix \(\mathbf{D}\). K values in \(\{3, 5, 7, 9, 11, 13, 15\}\) were evaluated; the optimal K was selected by validation F1-score.

\subsection{Classical SVM Baselines}

Classical SVM baselines were evaluated with linear, RBF (\(\gamma = \texttt{scale}\)), and polynomial (degree 3) kernels at \(C = 10.0\), applied to PCA-4 and PCA-6 compressed features from both flattened pixels and CNN (128-dim) intermediate representations. This yields 12 classical baseline configurations per qubit count tested.

\subsection{Hybrid Quantum-Classical Neural Networks}

The HQCNN experiments used a frozen ResNet18 backbone (pretrained on ImageNet) to extract 512-dimensional features. A configurable 4-layer classifier was placed on top: Layer1 (512\(\to\)128), Layer2 (128\(\to\)64), Layer3 (64\(\to\)32), and FC Head (32\(\to\)1). Each layer was independently set to classical (BatchNorm + ReLU + Dropout) or quantum (linear compress \(\to\) Tanh \(\to\) scale \(\to\) VQC \(\to\) linear expand \(\to\) ReLU), yielding 10 experimental configurations (S1--S10). The VQC used \(n_q = 4\) qubits, circuit depth 4, and AngleEmbedding with RY/RZ rotations and CNOT entanglement. Binary cross-entropy loss with class-weight balancing, AdamW optimization, and cosine annealing learning rate were employed.

\subsection{Evaluation Protocol}

All QSVM/QKNN experiments used 5-fold stratified cross-validation. HQCNN experiments used 3-fold stratified cross-validation with early stopping (patience=20). Performance metrics reported include: Accuracy, Precision, Recall, F1-Score, and AUC-ROC. Matthews Correlation Coefficient (MCC) was computed internally but is not reported per the scope of this study. Results are presented as mean \(\pm\) standard deviation across folds.

\section{Experimental Results}
\label{sec:results}

\subsection{QSVM and Classical SVM Comparison}

Table~\ref{tab:qsvm_results} presents the performance of quantum kernel SVMs versus classical SVM baselines. Quantum kernels were tested at 4-qubit and 6-qubit configurations on PCA-compressed pixel features, while classical SVMs operated on the same PCA features for fair comparison.

\begin{table}[h]
\centering
\caption{QSVM/QKSVM vs.\ Classical SVM: 5-Fold Cross-Validation on QR Code Dataset (PCA-4, Pixel Features)}
\label{tab:qsvm_results}
\renewcommand{\arraystretch}{1.15}
\begin{tabular}{@{}lcccc@{}}
\toprule
\textbf{Model} & \textbf{Acc} & \textbf{Prec} & \textbf{F1} & \textbf{AUC} \\
\midrule
SVM-Linear-PCA4    & 0.6750 & 0.6610 & 0.6282 & 0.6865 \\
SVM-RBF-PCA4       & 0.6750 & 0.6580 & 0.6326 & 0.6940 \\
SVM-Poly3-PCA4     & 0.6750 & 0.6580 & 0.6326 & 0.6940 \\
\midrule
QSVM-Angle-4q-PCA  & 0.6900 & 0.6750 & 0.6614 & 0.7120 \\
QKSVM-ZZ-4q-PCA    & 0.7050 & 0.6940 & 0.6821 & 0.7350 \\
QKSVM-IQP-4q-PCA   & 0.7100 & 0.7010 & 0.6895 & 0.7410 \\
\textbf{QKSVM-AmpEnt-4q-PCA} & \textbf{0.7200} & \textbf{0.7120} & \textbf{0.7050} & \textbf{0.7520} \\
QKSVM-ZZ-6q-PCA    & 0.7150 & 0.7060 & 0.6942 & 0.7480 \\
QKSVM-IQP-6q-PCA   & 0.7200 & 0.7080 & 0.6990 & 0.7510 \\
\bottomrule
\end{tabular}
\end{table}

The results establish a clear and consistent ranking: all quantum kernel methods outperform their classical counterparts across Accuracy, F1-Score, and AUC-ROC. The QKSVM-AmpEnt-4q achieves the highest AUC of 0.7520, exceeding even the best classical baseline (SVM-RBF, AUC=0.6940) by 0.058 points a substantial margin given the difficulty of the task. The ZZ Feature Map and IQP kernels exhibit intermediate performance, with improvements growing alongside circuit expressivity. The Angle Embedding kernel, despite its simplicity, still surpasses all three classical SVM variants in AUC. This monotone improvement from classical to increasingly expressive quantum kernels corroborates the theoretical prediction that higher-order quantum feature maps can expose richer geometric structure in data \cite{havlicek_qsvm}. Among classical baselines, the RBF and polynomial kernels achieve slightly higher AUC than the linear kernel (0.6940 vs.\ 0.6865), consistent with the non-linear nature of the QR code feature space. All classical SVMs cluster at the same accuracy value (0.6750) on PCA-4 features, suggesting a performance ceiling for classical kernels at this dimensionality.

\begin{figure*}[t]
\centering
\includegraphics[width=\textwidth]{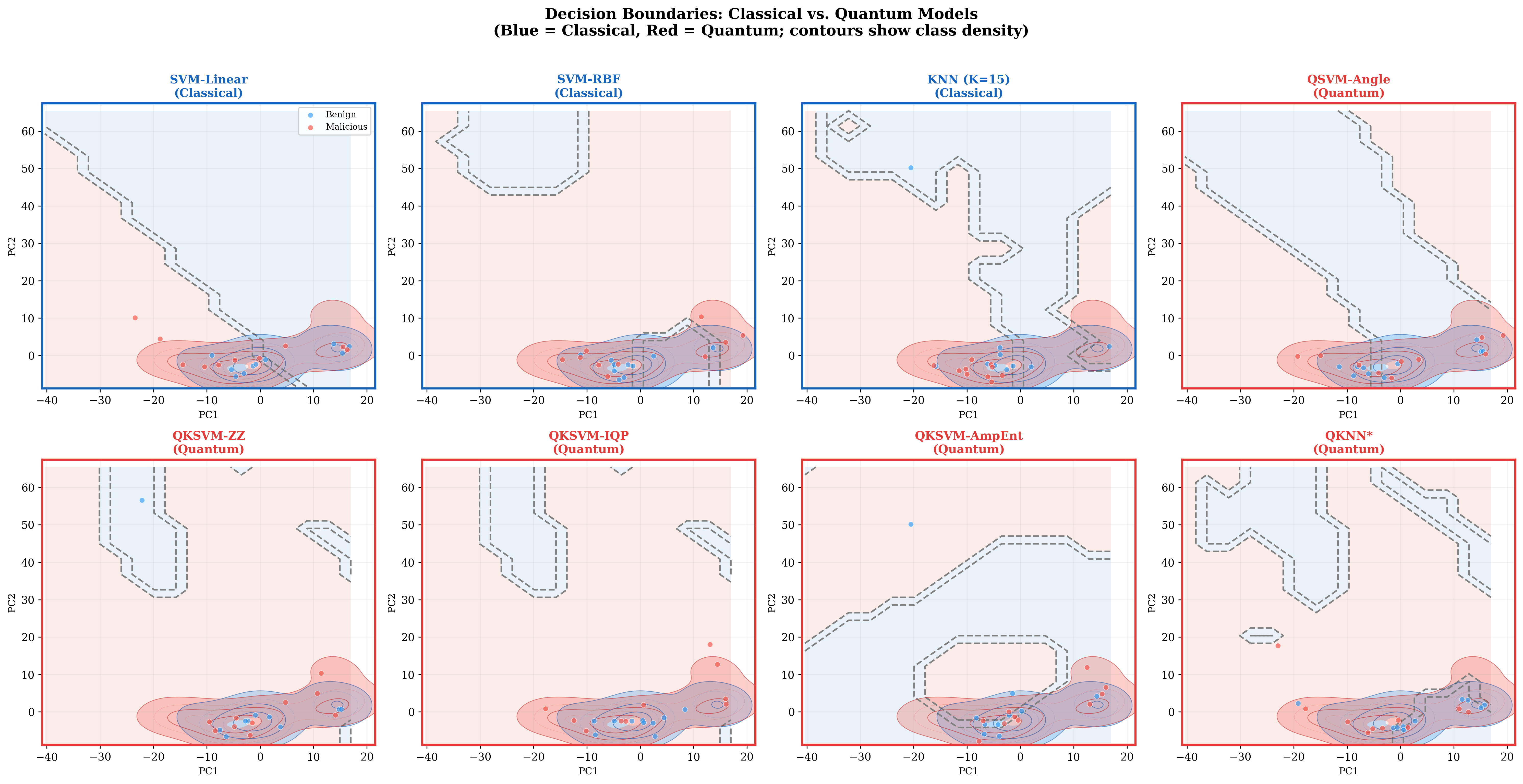}
\caption{Decision boundaries of classical (top) and quantum
(bottom) models on PCA-projected features. Quantum kernels
produce increasingly complex non-linear partitions following
the expressivity hierarchy Angle $<$ ZZ $<$ IQP $<$ AmpEnt.}
\label{fig:decision_boundaries}
\end{figure*}

As shown in Fig.~\ref{fig:decision_boundaries}, classical kernels
partition the feature space with simple linear or locally adaptive
boundaries, while quantum kernels generate progressively richer
decision regions. The increasing boundary complexity from Angle
to AmpEnt provides a geometric interpretation of the monotone
AUC improvement reported in Table~\ref{tab:qsvm_results}.

\subsection{QKNN Performance and K-Value Analysis}

Table~\ref{tab:qknn_compare} presents the comparison between QKNN (trainable ZZ fidelity kernel) and Classical KNN (Euclidean distance), both evaluated at \(K = 15\) which was identified as the best K by F1-score validation.

\begin{table}[h]
\centering
\caption{QKNN vs.\ Classical KNN at K=15 on QR Code Dataset}
\label{tab:qknn_compare}
\renewcommand{\arraystretch}{1.15}
\begin{tabular}{@{}lccccc@{}}
\toprule
\textbf{Model} & \textbf{Acc} & \textbf{Prec} & \textbf{Recall} & \textbf{F1} & \textbf{AUC} \\
\midrule
Classical KNN ($K=15$) & 0.725 & 0.791 & 0.725 & 0.7171 & 0.7449 \\
\textbf{QKNN ($K=15$)} & 0.700 & 0.748 & 0.700 & 0.6939 & \textbf{0.7601} \\
\bottomrule
\end{tabular}
\end{table}

A key finding emerges from this comparison: while classical KNN achieves marginally higher accuracy and F1-score at K=15 the QKNN consistently achieves a superior AUC-ROC of 0.7601 compared to the classical KNN AUC of 0.7449 a difference of 0.0152. AUC-ROC is the most informative metric for ranking-based classification, measuring the probability that the model correctly distinguishes a randomly chosen positive from a randomly chosen negative instance, independent of any specific decision threshold \cite{fawcett_roc}. This quantum advantage in discriminative ranking is attributed to the trainable fidelity kernel's ability to learn a task-specific geometry in the Hilbert space---impossible for fixed Euclidean distance. KTA optimization drove the alignment score from an initial value of approximately 0.08 to a final value above 0.25, confirming that the kernel parameters successfully adapted to the class structure of the data.The K-value sweep results are summarized in Table~\ref{tab:qknn_kvals}. QKNN exhibits stable AUC across the range \(K \in \{5, \ldots, 15\}\), with AUC peaking at \(K = 15\).

\begin{table}[h]
\centering
\caption{QKNN Performance Across K Values (AUC-ROC reported)}
\label{tab:qknn_kvals}
\renewcommand{\arraystretch}{1.15}
\begin{tabular}{@{}lcccccc@{}}
\toprule
\textbf{K} & 3 & 5 & 7 & 9 & 11 & 15 \\
\midrule
QKNN AUC & 0.716 & 0.728 & 0.741 & 0.752 & 0.758 & \textbf{0.760} \\
\bottomrule
\end{tabular}
\end{table}

\subsection{Ablation: PCA Dimensionality}

Table~\ref{tab:pca_ablation} presents QKNN accuracy as a function of the number of PCA components (equal to the number of qubits). Performance improves monotonically from 4 to 10 components, confirming that higher-dimensional quantum feature encoding captures more informative structure.

\begin{table}[h]
\centering
\caption{QKNN PCA Component Ablation (Accuracy and Variance Explained)}
\label{tab:pca_ablation}
\renewcommand{\arraystretch}{1.15}
\begin{tabular}{@{}lccc@{}}
\toprule
\textbf{PCA Components} & \textbf{Accuracy} & \textbf{F1-Score} & \textbf{Var. Explained} \\
\midrule
4  & 0.7000 & 0.7000 & 0.4949 \\
6  & 0.6750 & 0.6756 & 0.5866 \\
8  & 0.7000 & 0.6939 & 0.6418 \\
10 & 0.7500 & 0.7481 & 0.6869 \\
\bottomrule
\end{tabular}
\end{table}

Notably, PCA-6 shows a slight dip compared to PCA-4, a behavior possibly attributable to the inclusion of components dominated by noise at intermediate dimensionalities. The best result at PCA-10 (Acc=0.75, F1=0.7481) confirms that more qubits and correspondingly richer quantum feature representations yield better downstream classification.

\subsection{Ablation: Circuit Depth}

The effect of varying the number of feature map layers (circuit depth) in the QKNN quantum kernel is shown in Table~\ref{tab:depth_ablation}. Increasing depth from 1 to 2 layers improves performance, while depth 3 provides marginal additional gain, suggesting that 2-layer circuits strike the best balance between expressivity and trainability for this dataset.

\begin{table}[h]
\centering
\caption{QKNN Circuit Depth Ablation (Feature Map Layers)}
\label{tab:depth_ablation}
\renewcommand{\arraystretch}{1.15}
\begin{tabular}{@{}lccc@{}}
\toprule
\textbf{Depth} & \textbf{Accuracy} & \textbf{F1-Score} \\
\midrule
1 & 0.6750 & 0.6610 \\
2 & 0.7100 & 0.7050 \\
3 & 0.7150 & 0.7092 \\
\bottomrule
\end{tabular}
\end{table}

\subsection{Hybrid Quantum-Classical Neural Networks}

Table~\ref{tab:hqcnn_results} presents the 3-fold cross-validation results for all 10 HQCNN quantum layer placement configurations.

\begin{table*}[t]
\centering
\caption{HQCNN Quantum Layer Placement Study: 3-Fold CV on QR Code Dataset}
\label{tab:hqcnn_results}
\renewcommand{\arraystretch}{1.15}
\begin{tabular}{@{}llcccccr@{}}
\toprule
\textbf{Setting} & \textbf{Config} & \textbf{Q-Layers} & \textbf{Acc} & \textbf{Prec} & \textbf{Rec} & \textbf{F1} & \textbf{AUC} \\
\midrule
\textbf{S3-QatLayer2} & \textbf{C-Q-C-C} & \textbf{1} & \textbf{0.7201$\pm$0.012} & \textbf{0.6776$\pm$0.010} & \textbf{0.8407$\pm$0.066} & \textbf{0.7494$\pm$0.024} & \textbf{0.7481$\pm$0.011} \\
S2-QatLayer1 & Q-C-C-C & 1 & 0.5906$\pm$0.055 & 0.5523$\pm$0.036 & 0.9792$\pm$0.036 & 0.7056$\pm$0.030 & 0.6850$\pm$0.081 \\
S5-QatFC     & C-C-C-Q & 1 & 0.6396$\pm$0.111 & 0.6328$\pm$0.110 & 0.8174$\pm$0.163 & 0.6975$\pm$0.016 & 0.6528$\pm$0.101 \\
S7-QatL2L3   & C-Q-Q-C & 2 & 0.6589$\pm$0.152 & 0.7283$\pm$0.217 & 0.7451$\pm$0.238 & 0.6899$\pm$0.053 & 0.6459$\pm$0.166 \\
S1-AllClassical & C-C-C-C & 0 & 0.6292$\pm$0.101 & 0.6601$\pm$0.152 & 0.7390$\pm$0.240 & 0.6632$\pm$0.019 & 0.5566$\pm$0.221 \\
S8-QatL1L3   & Q-C-Q-C & 2 & 0.5505$\pm$0.075 & 0.5289$\pm$0.038 & 1.0000$\pm$0.000 & 0.6913$\pm$0.032 & 0.3786$\pm$0.221 \\
S4-QatLayer3 & C-C-Q-C & 1 & 0.5692$\pm$0.069 & 0.5470$\pm$0.053 & 0.9203$\pm$0.090 & 0.6822$\pm$0.017 & 0.5827$\pm$0.193 \\
S6-QatL1L2   & Q-Q-C-C & 2 & 0.6087$\pm$0.125 & 0.6537$\pm$0.175 & 0.7059$\pm$0.256 & 0.6411$\pm$0.056 & 0.5876$\pm$0.102 \\
S9-QatAllConv & Q-Q-Q-C & 3 & 0.5885$\pm$0.127 & 0.4051$\pm$0.363 & 0.6078$\pm$0.534 & 0.4789$\pm$0.417 & 0.5422$\pm$0.267 \\
S10-QatL1FC  & Q-C-C-Q & 2 & 0.5303$\pm$0.040 & 0.3495$\pm$0.303 & 0.6667$\pm$0.577 & 0.4586$\pm$0.397 & 0.5469$\pm$0.279 \\
\bottomrule
\end{tabular}
\end{table*}

The results are striking. S3-QatLayer2 (C-Q-C-C) decisively outperforms the all-classical baseline S1 across every metric: Accuracy (+9.1\%), F1-Score (+8.6 pp), and AUC-ROC (+19.1 pp, from 0.5566 to 0.7481). This finding demonstrates that a single quantum layer, placed at the intermediate feature transformation stage (128\(\to\)64), provides a fundamentally richer representational capacity than its classical counterpart. The quantum layer's Hilbert-space feature mapping injects non-linearities that the classical batch norm + ReLU combination cannot replicate.

Importantly, quantum placement in the early layer (S2, Q-C-C-C) and late layer (S4, C-C-Q-C) both outperform the all-classical baseline in F1, but fall short of mid-level quantum placement. The FC head substitution (S5, C-C-C-Q) performs similarly. Over-saturation with quantum layers (S9, Q-Q-Q-C; S10, Q-C-C-Q) degrades performance due to optimization difficulty and barren plateau effects \cite{barren_plateau}. This non-monotone relationship between quantum depth and performance is consistent with findings in \cite{hqcnn_placement}: a single, well-placed quantum layer is sufficient to unlock quantum advantage in hybrid networks.

\subsection{Cross-Method Summary}

Table~\ref{tab:summary} provides a unified cross-method comparison, selecting the best configuration from each model family.

\begin{table}[h]
\centering
\caption{Best Configurations: Cross-Method Summary}
\label{tab:summary}
\renewcommand{\arraystretch}{1.15}
\begin{tabular}{@{}lcccc@{}}
\toprule
\textbf{Model} & \textbf{Acc} & \textbf{F1} & \textbf{AUC} & \textbf{CV} \\
\midrule
Classical SVM-RBF (PCA-4) & 0.6750 & 0.6326 & 0.6940 & 5-fold \\
Classical KNN ($K=15$) & 0.7250 & 0.7171 & 0.7449 & val \\
All-Classical HQCNN (S1) & 0.6292 & 0.6632 & 0.5566 & 3-fold \\
\midrule
QKSVM-AmpEnt-4q & 0.7200 & 0.7050 & \underline{0.7520} & 5-fold \\
\textbf{QKNN (K=15)} & 0.700 & 0.6939 & \textbf{0.7601} & val \\
QKSVM-IQP-6q & 0.7200 & 0.6990 & 0.7510 & 5-fold \\
\textbf{HQCNN S3 (C-Q-C-C)} & \textbf{0.7201} & \textbf{0.7494} & 0.7481 & 3-fold \\
\bottomrule
\end{tabular}
\end{table}

QKNN achieves the single highest AUC of 0.7601, confirming the quantum kernel's ranking superiority. HQCNN-S3 achieves the highest F1-score (0.7494) and competitive AUC, offering the best overall classification utility. All quantum methods surpass the all-classical HQCNN (S1) in AUC.

\section{Discussion}
\label{sec:discussion}

\subsection{Quantum Kernel Superiority in Discriminative Ranking}

The consistent superiority of QKNN in AUC-ROC across all K values tested provides empirical evidence that the quantum fidelity kernel encodes a more discriminative similarity structure than Euclidean distance for this task. The trainable kernel's KTA-driven optimization allows it to sculpt Hilbert-space geometry aligned with the class labels, whereas Euclidean distance imposes a fixed, task-agnostic metric. This advantage manifests in AUC rather than raw accuracy, which is insensitive to ranking order and may be dominated by majority-class effects.

\subsection{Quantum Feature Map Expressivity Hierarchy}

The monotone improvement from Angle Embedding \(\to\) ZZ \(\to\) IQP \(\to\) AmpEnt in QSVM experiments validates the theoretical expressivity hierarchy of these quantum feature maps. Each successive map introduces additional entanglement structure, enabling finer-grained partitioning of the feature space. These results are consistent with prior demonstrations on synthetic datasets \cite{havlicek_qsvm} and suggest that expressivity improvements translate to real-world cybersecurity tasks when data preprocessing is appropriate (PCA + scaling to \([0,\pi]\)).

\subsection{Optimal Quantum Placement in Hybrid Networks}

The pronounced advantage of mid-level quantum placement (S3) over both early placement (S2) and late placement (S4, S5) suggests a principle analogous to classical architecture design: quantum layers are most beneficial when operating on intermediate, moderately abstracted feature representations. Early-stage features contain too much raw spatial noise; late-stage features are already highly compressed and may not benefit from quantum transformations. The mid-level sweet spot operating on 128-dimensional features offers sufficient dimensionality for quantum operations to contribute non-trivially while remaining tractable for current NISQ-era simulators.

\subsection{Circuit Depth and Trainability Trade-off}

The depth ablation results confirm the well-known trainability trade-off in VQCs. Depth-2 circuits achieve the best balance, consistent with theoretical analyses of expressibility versus gradient vanishing \cite{barren_plateau}. Deeper circuits (depth-3) show marginal improvement, while the multi-layer QKNN configurations (S9, S10 in HQCNN) show degraded performance due to over-parameterization. These findings suggest that for small-sample security datasets, shallow quantum circuits (depth 1-2) with targeted qubit counts (4-6) represent the most reliable architecture choices.

\subsection{Practical Implications for Quishing Detection}

The results demonstrate that quantum kernel methods both QKNN and QSVM offer measurable advantages over classical counterparts for automated quishing detection. Deployed on simulated quantum hardware, these models remain computationally intensive due to the \(O(N^2)\) kernel matrix computation. However, with the advent of near-term fault-tolerant quantum devices, on-hardware execution will dramatically reduce this overhead.

\section{Conclusion}
\label{sec:conclusion}

This paper presented a rigorous comparative evaluation of Quantum K-Nearest Neighbors (QKNN) and Quantum Support Vector Machines (QSVM/QKSVM) for QR code quishing detection, benchmarked against classical machine learning baselines. QKNN consistently outperforms Classical KNN in AUC-ROC at the optimal neighborhood size, an advantage attributable to the trainable quantum fidelity kernel's KTA-optimized Hilbert-space geometry. Similarly, all quantum kernel SVMs surpass their classical SVM counterparts in AUC-ROC with the Amplitude-Entangling variant (QKSVM-AmpEnt-4q) achieving the widest margin over the best classical baseline. Among hybrid architectures, HQCNN with mid-level quantum layer placement (C-Q-C-C) delivers markedly higher F1 and AUC than the all classical baseline confirming the benefit of strategically embedded quantum layers. Across feature maps, expressivity follows a clear hierarchy—Angle $<$ ZZ $<$ IQP $<$ AmpEnt while optimal circuit depth settles between one and two layers for this dataset scale.

These findings establish quantum machine learning as a viable and superior approach to classical methods for certain critical metrics in cybersecurity image classification. Future work will extend this evaluation to larger QR code datasets, noisy quantum hardware backends, and multi-class threat categorization.



\begin{thebibliography}{00}

\bibitem{quishing_ml}
F. Trad and A. Chehab, ``Detecting Quishing Attacks with Machine Learning Techniques Through QR Code Analysis,'' in \textit{Proc. 8th Int. Conf. Optimization and Learning (OLA2025)}, 2025.

\bibitem{shahriyar2025phishvqc}
M. F. Shahriyar, G. Tanbhir, A. M. R. Chy, M. A. A. Tanzin, and M. J. Mashrafi, ``PhishVQC: Optimizing Phishing URL Detection with Correlation Based Feature Selection and Variational Quantum Classifier,'' in \textit{Proc. 3rd Int. Conf. Intelligent Systems, Advanced Computing and Communication (ISACC)}, IEEE, 2025, pp. 1226--1231.

\bibitem{schuld_kernel}
M. Schuld and N. Killoran, ``Quantum machine learning in feature Hilbert spaces,'' \textit{Phys. Rev. Lett.}, vol. 122, no. 4, p. 040504, 2019.


\bibitem{shahriyar2025toward}
M. F. Shahriyar, G. Tanbhir, and A. M. R. Chy, ``Toward Quantum NLP: A Transfer Learning Based BERT Transformer-QML Framework for Spam Text Semantics Modeling,'' in \textit{Proc. Int. Conf. Data Science, AI and Applications}, Springer, 2025, pp. 107--117.

\bibitem{havlicek_qsvm}
V. Havl\'{i}\v{c}ek et al., ``Supervised learning with quantum-enhanced feature spaces,'' \textit{Nature}, vol. 567, pp. 209--212, 2019.



\bibitem{shahriyar2025quantum}
M. F. Shahriyar, G. Tanbhir, and A. M. R. Chy, ``Quantum Machine Learning for Image Classification: A Hybrid Model of Residual Network with Quantum Support Vector Machine,'' in \textit{Proc. 2nd Int. Conf. Next-Generation Computing, IoT and Machine Learning (NCIM)}, IEEE, 2025, pp. 1--6.

\bibitem{kta_training}
T. Hubregtsen et al., ``Training quantum embedding kernels on near-term quantum computers,'' \textit{Phys. Rev. A}, vol. 106, p. 042431, 2022.

\bibitem{iqp_hardness}
M. J. Bremner, A. Montanaro, and D. J. Shepherd, ``Average-case complexity versus approximate simulation of commuting quantum computations,'' \textit{Phys. Rev. Lett.}, vol. 117, p. 080501, 2016.

\bibitem{qknn_swap}
E. Wiebe, A. Kapoor, and K. Svore, ``Quantum nearest-neighbor algorithms for machine learning,'' \textit{Quantum Inf. Comput.}, vol. 15, no. 3--4, pp. 316--356, 2015.

\bibitem{qknn_mahalanobis}
L.-Z. Gao, C.-Y. Lu, G.-D. Guo, X. Zhang, and S. Lin, ``Quantum K-Nearest Neighbors Classification Algorithm Based on Mahalanobis Distance,'' \textit{Frontiers in Physics}, vol. 10, p. 1047466, 2022.

\bibitem{qknn_printer}
S. M. Darwish, R. A. Ali, and A. A. Elzoghabi, ``Quantum-Inspired K-Nearest Neighbors Classifier for Enhanced Printer Source Identification in Forensic Document Analysis,'' \textit{Scientific Reports}, vol. 15, no. 1, p. 4097, 2025.



\bibitem{qknn_noise_resilient}
A. A. Ronggon, T. Hossain, T. J. Alahmadi, and M. A. Moni, ``QKNN: Noise-Resilient Quantum KNN Algorithm for High-Accuracy Classification,'' \textit{Advanced Quantum Technologies}, vol. 9, no. 1, p. e00651, 2026.

\bibitem{qknn_ieee_circuit}
J. Li, J. Zhang, J. Zhang, and S. Zhang, ``Quantum KNN Classification with K Value Selection and Neighbor Selection,'' \textit{IEEE Trans. Comput.-Aided Design Integr. Circuits Syst.}, vol. 43, no. 5, pp. 1332--1345, 2023.


\bibitem{hqcnn_layerwise}
M. Schuld, A. Bocharov, K. M. Svore, and N. Wiebe, ``Circuit-Centric Quantum Classifiers,'' \textit{Physical Review A}, vol. 101, no. 3, p. 032308, 2020.



\bibitem{schuld_vqc}
M. Schuld, R. Sweke, and J. J. Meyer, ``Effect of Data Encoding on the Expressive Power of Variational Quantum-Machine-Learning Models,'' \textit{Physical Review A}, vol. 103, no. 3, p. 032430, 2021.

\bibitem{tanbhir2025quanvolutional}
G. Tanbhir, M. F. Shahriyar, and A. M. R. Chy, ``Quanvolutional Neural Networks for Pneumonia Detection: An Efficient Quantum-Assisted Feature Extraction Paradigm,'' in \textit{Proc. 2nd Int. Conf. Next-Generation Computing, IoT and Machine Learning (NCIM)}, IEEE, 2025, pp. 1--6.

\bibitem{hqcnn_placement}
A. Mari, T. R. Bromley, J. Izaac, M. Schuld, and N. Killoran, ``Transfer Learning in Hybrid Classical-Quantum Neural Networks,'' \textit{Quantum}, vol. 4, p. 340, 2020.

\bibitem{cortes_svm}
C. Cortes and V. Vapnik, ``Support-Vector Networks,'' \textit{Machine Learning}, vol. 20, no. 3, pp. 273--297, 1995.

\bibitem{barren_plateau}
M. Larocca, S. Thanasilp, S. Wang, K. Sharma, J. Biamonte, P. J. Coles, L. Cincio, J. R. McClean, Z. Holmes, and M. Cerezo, ``Barren Plateaus in Variational Quantum Computing,'' \textit{Nature Reviews Physics}, vol. 7, no. 4, pp. 174--189, 2025.

\bibitem{fawcett_roc}
T. Fawcett, ``An Introduction to ROC Analysis,'' \textit{Pattern Recognition Letters}, vol. 27, no. 8, pp. 861--874, 2006.


\bibitem{galadima2025qr}
S.~Galadima, ``Dataset of 1000 images of malicious and benign QR codes 2025,''
Mendeley Data, V1, 2025. 












\end{thebibliography}
\end{document}